\documentclass[runningheads]{llncs}
\newcommand{\corrauth}{\textsuperscript{(\Letter)}}
\usepackage{graphicx}
\usepackage{booktabs}

\usepackage{xcolor}

\usepackage{multirow}
\usepackage{cite}
\usepackage[misc]{ifsym}
\usepackage{algorithm}
\usepackage{algpseudocode}

\begin{document}
\title{Soft-tissue Driven Craniomaxillofacial Surgical Planning}

\author{Xi Fang\inst{1} \and
Daeseung Kim\inst{2}\corrauth \and
Xuanang Xu\inst{1} \and
Tianshu Kuang\inst{2} \and
Nathan~Lampen\inst{1} \and
Jungwook Lee\inst{1} \and
Hannah H. Deng\inst{2} \and
Jaime Gateno\inst{2} \and
Michael~A.K.~Liebschner\inst{3} \and
James J. Xia\inst{2} \and
Pingkun~Yan\inst{1}\corrauth}
\renewcommand{\thefootnote}{\fnsymbol{footnote}}
\footnotetext[2]{D. Kim (\email{DKim@houstonmethodist.org}) and P. Yan (\email{yanp2@rpi.edu}) are co-corresponding authors.}
\authorrunning{X.~Fang et al.}
\institute{Department of Biomedical Engineering and Center for Biotechnology and Interdisciplinary Studies, Rensselaer Polytechnic Institute, Troy, NY 12180, USA
\and 
Department of Oral and Maxillofacial Surgery, Houston Methodist Research Institute, Houston, TX 77030, USA
\and 
Department of Neurosurgery, Baylor College of Medicine, Houston, TX 77030, USA}
\maketitle              
\begin{abstract}
In CMF surgery, the planning of bony movement to achieve a desired facial outcome is a challenging task.
Current bone driven approaches focus on normalizing the bone with the expectation that the facial appearance will be corrected accordingly.
However, due to the complex non-linear relationship between bony structure and facial soft-tissue, such bone-driven methods are insufficient to correct facial deformities. 
Despite efforts to simulate facial changes resulting from bony movement, surgical planning still relies on iterative revisions and educated guesses.
To address these issues, we propose a soft-tissue driven framework that can automatically create and verify surgical plans.
Our framework consists of a bony planner network that estimates the bony movements required to achieve the desired facial outcome and a facial simulator network that can simulate the possible facial changes resulting from the estimated bony movement plans.
By combining these two models, we can verify and determine the final bony movement required for planning.
The proposed framework was evaluated using a clinical dataset, and our experimental results demonstrate that the soft-tissue driven approach greatly improves the accuracy and efficacy of surgical planning when compared to the conventional bone-driven approach.

\end{abstract}
\section{Introduction}

Craniomaxillofacial (CMF)  deformities can affect the skull, jaws, and midface. 
When the primary cause of disfigurement lies in the skeleton, surgeons cut the bones into pieces and reposition them to restore normal alignment \cite{shafi2013accuracy}.  In this context, the focus is on correcting bone deformities, as it is anticipated that the restoration of normal facial appearance will follow automatically. Consequently, it is customary to initiate the surgical planning process by estimating the positions of normal bones. The latter has given rise to bone-driven approaches \cite{bobek2015virtual, mccormick2011virtual, xia2015algorithm}.  For example, methods based on sparse representation \cite{wang2015estimating} and deep learning \cite{xiao2021estimating} have been proposed to estimate the bony shape that may lead to an acceptable facial appearance. 

However, the current bone-driven methods have a major limitation, subjecting to the complex and nonlinear relationship between the bones and the draping soft-tissues. Surgeons estimate the required bony movement through trial and error, while computer-aided surgical simulation (CASS) software \cite{xia2015algorithm} simulates the effect on facial tissues resulting from the proposed movements. Correcting the bone deformity may not completely address the soft-tissue disfigurement. The problem can be mitigated by iteratively revising the bony movement plan and simulating the corresponding soft-tissue changes.
However, this iterative planning revision is inherently time-consuming, especially when the facial change simulation is performed using computationally expensive techniques such as the finite-element method (FEM) \cite{kim2021novel}.
Efforts have been made to accelerate facial change prediction using deep learning algorithms \cite{ma2021deep,fang2022deep,lampen2022deep}, which, however,  do not change the iterative nature of the bone-driven approaches.

To address the above challenge, this paper proposes a novel soft-tissue driven surgical planning framework. Instead of simulating facial tissue under the guessed movement, our approach directly aims at a desired facial appearance and then determines the optimal bony movements required to achieve such an appearance without the need for iterative revisions.
Unlike the bone-driven methods, this soft-tissue driven framework eliminates the need for surgeons to make educated guesses about bony movement, significantly improving the efficiency of the surgical planning process. 
Our proposed framework consists of two main components, the Bony Planner (BP) and the Facial Simulator (FS). The BP estimates the possible bony movement plans (bony plans) required to achieve the desired facial appearance change, while the FS verifies the effectiveness of the estimated plans by simulating the facial appearance based on the bony plans.
Without the intervention of clinicians, the BP automatically creates the most clinically feasible surgical plan that achieves the desired facial appearance.

The main contributions of our work are as follows. 1) This is the first soft-tissue driven approach for CMF surgical planning, which can substantially reduce the planning time by removing the need for repetitive guessing bony movement.
2) We develop a deep learning model as the bony planner, which can estimate the underlying bony movement needed for changing a facial appearance into a targeted one.
3) The developed FS module can qualitatively assess the effect of surgical plans on facial appearance, for virtual validation.

\section{Method}
\begin{figure}[t]
	\centering
	\includegraphics[width = \textwidth] {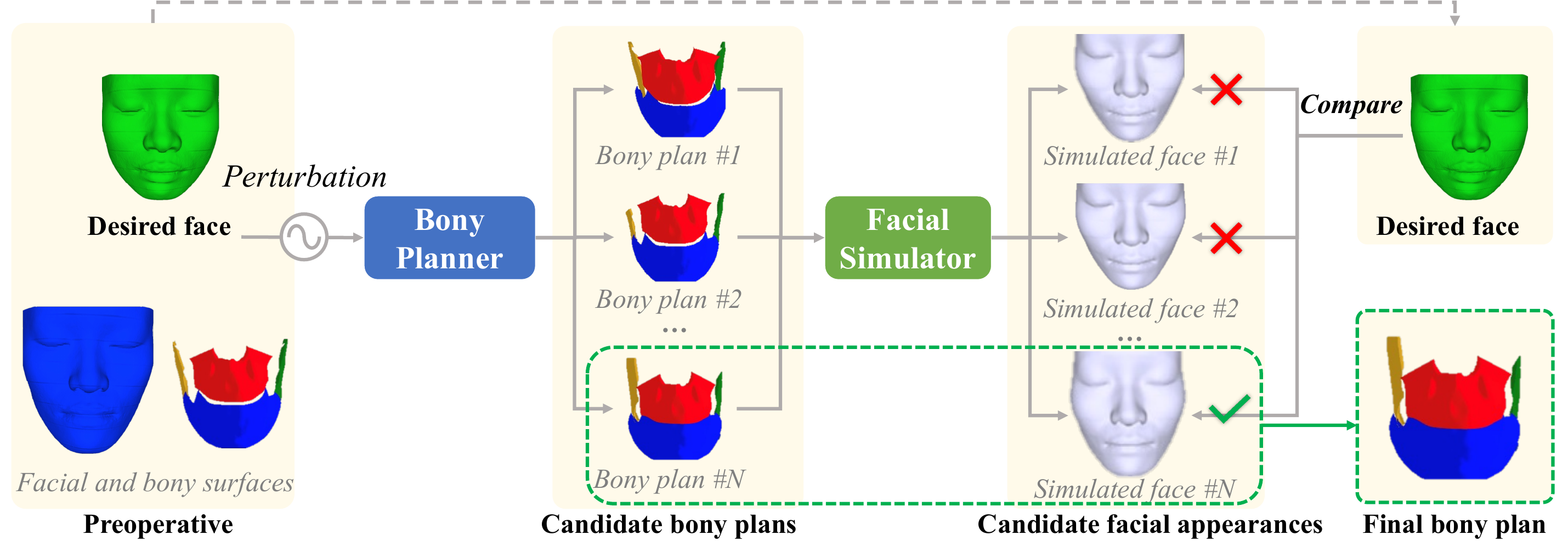}
	\caption{Overview of the proposed soft-tissue driven framework. The framework is composed of two main components: the creation of candidate bony plans using BP, and the simulation of facial outcomes following the plans using FS. Finally, the facial outcomes are compared with the desired face to select the final bony plan.
	}
	\label{fig:overview}
\end{figure}

Fig.~\ref{fig:overview} shows an overview of the proposed framework, which consists of two primary modules. The first module is a BP network that plans bony movement based on the desired facial outcome and the given preoperative facial and bony surfaces. The second module is the FS network that simulates corresponding postoperative facial appearances by applying the estimated bony plans. 
Instead of providing one single bony plan, the BP will estimate multiple plans because not all the generated plans may result in the desired clinical effect. The FS then simulates facial appearances by using those plans and chooses the plan leading to the facial appearance closest to the desired target. The two models are deployed together to determine and confirm the final bony plan. Below we first present the details of BP and FS modules, then we introduce how they are deployed together for inference.

\subsection{Bony Planner (BP)}
\begin{figure}[t]
	\centering
	\includegraphics[width = \textwidth] {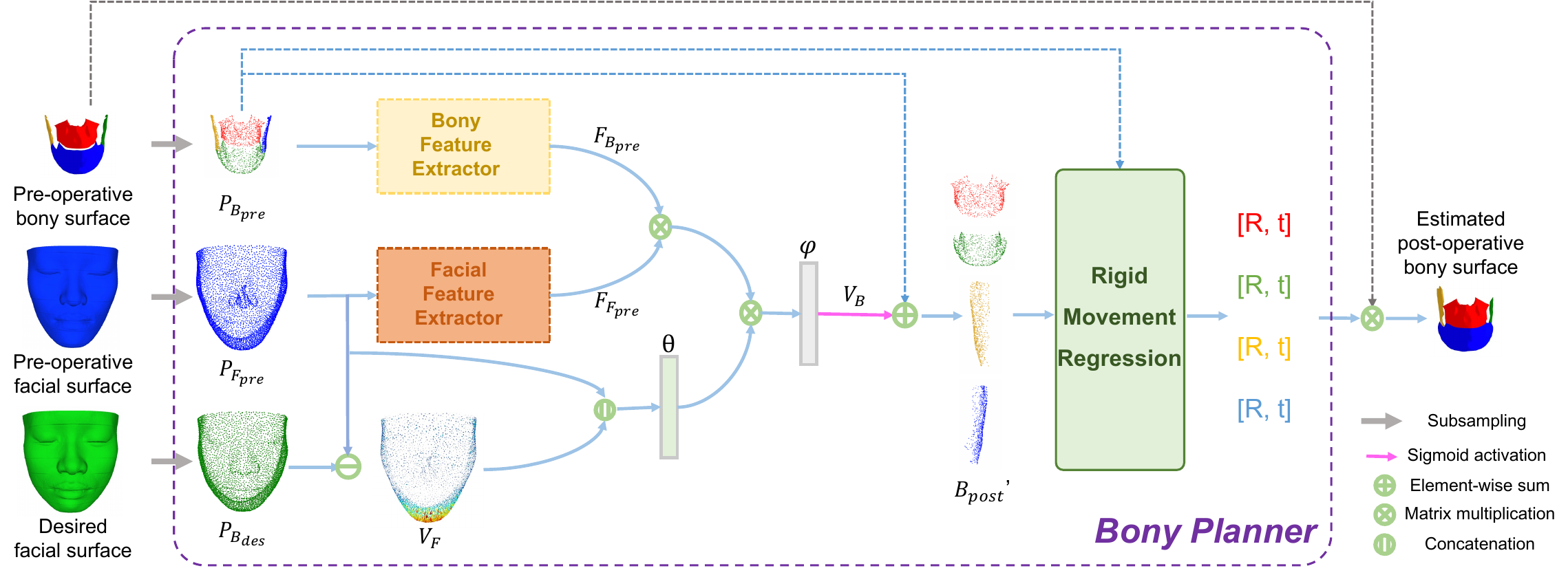}
	\caption{Scheme of the proposed bony planner (BP). BP first estimates the non-rigidly deformed bony points based on the desired face, then regresses the rigid movements by fitting the non-rigid prediction.
	}
 \label{fig:bp}
\end{figure}

\textbf{Data preprocessing:} 
The bony plan is created by the BP network using preoperative facial $F_{pre}$, bony surface $B_{pre}$, and desired facial surface $F_{des}$. The goal of the BP network is to convert the desired facial change from $F_{pre}$ to $F_{des}$ into rigid bony movements, denoted as $T_{S}$ for each bony segment $S$, as required for surgical planning. However, it is very challenging to directly estimate the rigid bone transformation from the facial difference. Therefore, we first estimate the non-rigid bony movement vector field and then convert that into the rigid transformations for each bone segment. Fig~\ref{fig:bp} illustrates the BP module and the details are provided as follows. For computational efficiency, pre-facial point set $P_{F_{pre}}$, pre-bony point set $P_{B_{pre}}$, desired point set $P_{F_{des}}$ are subsampled from the pre-facial/bony and desired facial surfaces.
\\\\
\textbf{Non-rigid bony movement vector estimation:}
We adopt the Attentive Correspondence assisted Movement Transformation network (ACMT-Net) \cite{fang2022deep}, which was originally proposed for facial tissue simulation, to estimate the point-wise bony displacements to acquire $F_{des}$. 
This method is capable of effectively computing the movement of individual bony points by capturing the relationship between facial points through learned affinity.

In the network, point-wise facial features ($F_{F_{pre}}$) and bony features ($F_{B_{pre}}$) are extracted from $P_{F_{pre}}$ and $P_{B_{pre}}$. We used a pair of modified PointNet++ modules where the classification layers were removed and a 1D convolution layer was added at the end to project $F_{F_{pre}}$ and $F_{B_{pre}}$ into the same lower dimensions \cite{qi2017pointnet++}. 
A correlation matrix $R$ is established by computing the normalized dot product between $F_{F_{pre}}$ and $F_{B_{pre}}$ to evaluate the relationship between the facial and bony surfaces: 
\begin{equation}
R = \frac{{F_{B_{pre}}}^T{F_{F_{pre}}}}{N_{F_{pre}}},
\end{equation}
where $N_{F_{pre}}$ denotes the number of facial points $P_{F_{pre}}$.
On the other hand, desired facial movement $V_F$ is computed by subtracting $P_{F_{des}}$ and $P_{F_{pre}}$.
$V_F$ is then concatenated with $P_{F_{pre}}$ and fed into a 1D convolution layer to encode the movement information.
Then the movement feature of each bony point is estimtated by the normalized summary of facial features using $R$.
Finally, the transformed bony movement features are decoded into movement vectors after being fed into one 1D convolution layer and normalization:
\begin{equation}
V_B = \varphi({\theta([P_{F_{pre}}, V_F]}R).
\end{equation}\\
\textbf{Rigid bony movement regression:}
Upon obtaining the estimated point-wise bony movements, they are added to the corresponding bony points, resulting in a non-rigidly transformed bony point set denoted as $P_{B_{pdt}}$. The resulting points are grouped based on their respective bony segments, with point sets $P_{S_{pre}}$ and $P_{S_{pdt}}$ representing each bony segment $S$ before and after movement, respectively.
To fit the movement between $P_{S_{pre}}$ and $P_{S_{pdt}}$, we estimate the rigid transformations $[R_{S}, T_{S}]$ by minimizing the mean square error as follows:
\begin{equation}
    E(R_{S}, T_{S}) = \big\|\frac{1}{N}\sum_{i}^{n}{(R_{S}P_{S_{pre}}^{i} + T_{S} - P_{S_{pdt}}^{i})}\big\|^2,
\end{equation}
where $i$ and $N$ represent the $i$-th point and the total number of points in $P_{S_{pre}}$, respectively. 
First, we define the centroids of $P_{S_{pre}}$ and $P_{S_{pdt}}$ to be $\overline{P_{S_{pre}}}$ and $\overline{P_{S_{pdt}}}$.
The cross-covariance matrix $\mathbf{H}$ can be computed as 
\begin{equation}
    \mathbf{H} = \sum_{i=1}^N{(P_{S_{pre}}-\overline{P_{S_{pre}}})(P_{S_{pdt}}-\overline{P_{S_{pdt}}})}
\end{equation}
Then we can use singular value decomposition (SVD) to decompose 
\begin{equation}
    \mathbf{H} = USV^T,
\end{equation}
then the alignment minimizing $E(R_{S}, T_{S})$ can be solved by
\begin{equation}
    R_{S} = VU^T,
    T_{S} = -R_{S}\overline{P_{S_{pre}}} + \overline{P_{S_{pdt}}}.
\end{equation}
Finally, the rigid transformation matrices are applied to their corresponding bone segments for virtual planning.

\subsection{Facial Simulator (FS)}
While the bony planner can estimate bony plans and we can compare them with ground truth. However, the complex relationship between bone and face makes it unknown whether adopting the bony plan will result in the desired facial outcome or not. 
To evaluate the effectiveness of the BP network in simulating facial soft tissue, an FS is developed to simulate the facial outcome following the estimated bony plans. For facial simulation, we employ the ACMT-Net, which takes $P_{F_{pre}}$, $P_{B_{pre}}$, and $P_{B_{pdt}}$ as input and predicts the point-wise facial movement vector $V_F^{'}$.
The movement vector of all vertices in the facial surface is estimated by interpolating $V_F^{'}$. The simulated facial surface $F_{pdt}^{'}$ is then reconstructed by adding the predicted movement vectors to the vertices of $F_{pre}$.

\subsection{Self-verified virtual planning}
To generate a set of potential bony plans, we randomly perturbed the surfaces by flipping and translating them up to 10mm in three directions during inference. 
We repeated this process 10 times in our work. 
After estimation, the bony surfaces were re-localized to their original position prior to perturbation. 
Sequentially, the FS module generated a simulated facial appearance for each bony plan estimated from the BP module, serving two purposes. 
Firstly, it verified the feasibility of the bony plan through facial appearance. Secondly, it allowed us to evaluate the facial outcomes of different bony plans. 
The simulated facial surfaces were compared with the desired facial surface, and the final plan was selected based on the similarity of the resulting facial outcome. 
This process verified the efficacy of the selected plan for achieving the desired facial outcome.

\section{Experiments and Results}
\begin{figure}[t]
	\centering
	\includegraphics[width = \textwidth] {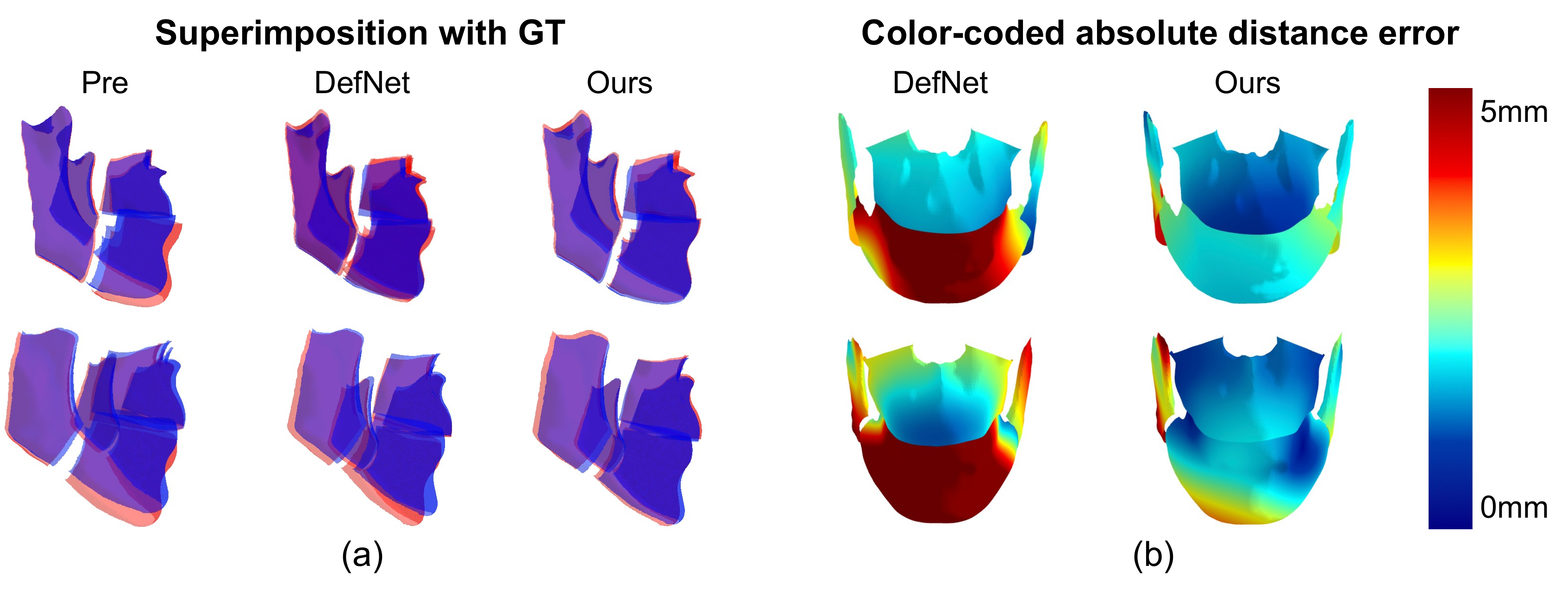}
	\caption{Examples of the results of bony plans. (a) Visual comparison of the estimated bony surface(red) with ground truth (blue), where deeper shades of blue and red indicate shading effects. (b) Color-coded error map to display the difference between the estimated bony surface and ground truth.
	}
	\label{fig:result}
\end{figure}
\subsection{Dataset}
We employed a five-fold cross-validation technique to evaluate the performance of the proposed network using 34 sets of patient CT data. 
We partitioned the data into five groups with \{7, 7, 7, 7, 6\} sets of data, respectively. During each round of validation, four of these groups (folds) were used for training and the remaining group was used for testing.
The CT scans were randomly selected from our digital archive of patients who had undergone double-jaw orthognathic surgery.
To obtain the necessary data, we employed a semi-automatic method to segment the facial and bony surface from the CT images \cite{liu2021skullengine}.
Then we transformed the segmentation into surface meshes using the Marching Cube approach \cite{lorensen1987marching}.
To retrospectively recreate the surgical plan that could “achieve” the actual postoperative outcomes, we registered the postoperative facial and bony surfaces to their respective preoperative surfaces based on surgically unaltered bony volumes, i.e., cranium, and utilized them as a roadmap \cite{ma2021deep}.
To establish the surgical plan to achieve actual postoperative outcomes, virtual osteotomies were first reperformed on the preoperative bones to create bony segments, including LeFort 1 (LF), distal (DI), right proximal (RP), and left proximal (LP). 
The movement of the bony segments, which represents the surgical plan to achieve the actual postoperative outcomes, was retrospectively established by manually registering each bony segment to the postoperative roadmap, which served as the ground truth in the evaluation process.
We rigidly registered the segmented bony and facial surfaces to their respective bony and facial templates to align different subjects. Additionally, we cropped the facial surfaces to retain only the regions of interest for CMF surgery. 
In this study, we assume the postoperative face is the desired face.
For efficient training, we subsampled 4096 points from the facial and bony surfaces, respectively, with 1024 points for each bony segment. 
To augment the data, we randomly flipped the point sets symmetrically and translated them along three directions within a range of 10mm.
\subsection{Implementation and evaluation methods}
To compare our approach, we implemented the state-of-the-art bone-driven approach, i.e., deformation network (DefNet) \cite{xiao2021estimating},  which takes the point coordinates and their normal vectors as input and generates the displacement vectors to deform the preoperative bones. 
Our method has two variations: BP and BP+FS. BP estimates only one plan, while BP+FS selects the final plan based on the simulated facial outcomes of our facial simulator. 
The PointNet++ networks utilized in both BP and FS are comprised of four feature-encoding blocks and four feature-decoding blocks.
The output dimensions for each block are 128, 256, 512, 1024, 512, 256, 128, and 128, respectively, and the output point numbers of the modules are 1024, 512, 256, 64, 256, 512, 1024, and 4096, sequentially.
We used the Adam optimizer with a learning rate of 0.001, Beta1 of 0.9, and Beta2 of 0.999 to train both the BP and FS networks. 
For data augmentation, both facial and bone data are subjected to the same random flipping and translation, ensuring that the relative position and scale of facial changes and bony movements remain unchanged.
The models were trained for 500 epochs with a batch size of 4 and MSE loss was used, after which the models were used for evaluation. 
The models were trained on an NVIDIA DGX-1 deep learning server with eight V100 GPUs. 

The prediction accuracy of the soft-tissue driven approach was evaluated quantitatively and qualitatively by comparing it with DefNet.
Then quantitative evaluation was to assess the accuracy using the mean absolute error (MAE) between the predicted bony surfaces and the ground truth.
For detailed evaluation, MAE was also separately calculated for each bony segment, including LF, DI, RP, and LP. 
Statistical significance was determined using the Wilcoxon signed-rank test to compare the results obtained by different methods \cite{rey2011wilcoxon}. 
Qualitative evaluation was carried out by directly comparing the bony surfaces and simulated facial outcomes generated by our approach and DefNet with the ground truth postoperative bony and facial surfaces.
\begin{table}[t]
\centering
\caption{Quantitative evaluation results. Prediction accuracy comparison with the state-of-the-art bone-driven methods.}
\begin{tabular}{ |l|c|c|c|c|c| }
	\hline
	\multirow{2}{6em}{Method} & \multicolumn{5}{|c|}{Mean absolute distance (mean $\pm$ std) in millimeter}\\
	\cline{2-6}
	& LF & DI & RP & LP & Entire Bone \\
	\hline
	DefNet \cite{xiao2021estimating} & $2.69\pm1.18$ & $6.45\pm1.79$ & $3.08\pm1.15$ &  $3.23\pm1.29$ & $3.86\pm0.91$ 	\\
	BP & $2.42\pm1.13$ & $3.16\pm1.38$ & $2.58\pm1.29$ & $2.42\pm1.16$  & $2.64\pm0.97$ 	\\
	BP + FS & $2.46\pm1.13$ & $2.89\pm1.64$ & $2.56\pm1.30$ &	$2.41\pm1.13$ & $2.58\pm0.93$ 	\\
	\hline
	\end{tabular}
 \label{tab:1}
\end{table}
\subsection{Results}
The results of the quantitative evaluation are shown in Table~\ref{tab:1}. The results of the Wilcoxon signed-rank test showed that BP outperforms DefNet on LF ($p<0.05$), DI ($p<0.001$), LP segments ($p<0.01$), and the entire bone ($p<0.001$) by statistically significant margins. 
In addition, BP+FS significantly outperforms BP on DI segment ($p<0.05$).
 Two randomly selected patients are shown in Fig.~\ref{fig:result}. 
 To make a clear visual comparison, we superimposed the estimated bony surfaces (Blue) with their corresponding ground truth (Red) and set the surfaces to be transparent as shown in Fig.~\ref{fig:result} (a).
 Fig.~\ref{fig:result} (b) displays the surface distance error between the estimated bony surfaces and ground truth.

\begin{figure}[t]
	\centering
	\includegraphics[width = \textwidth] {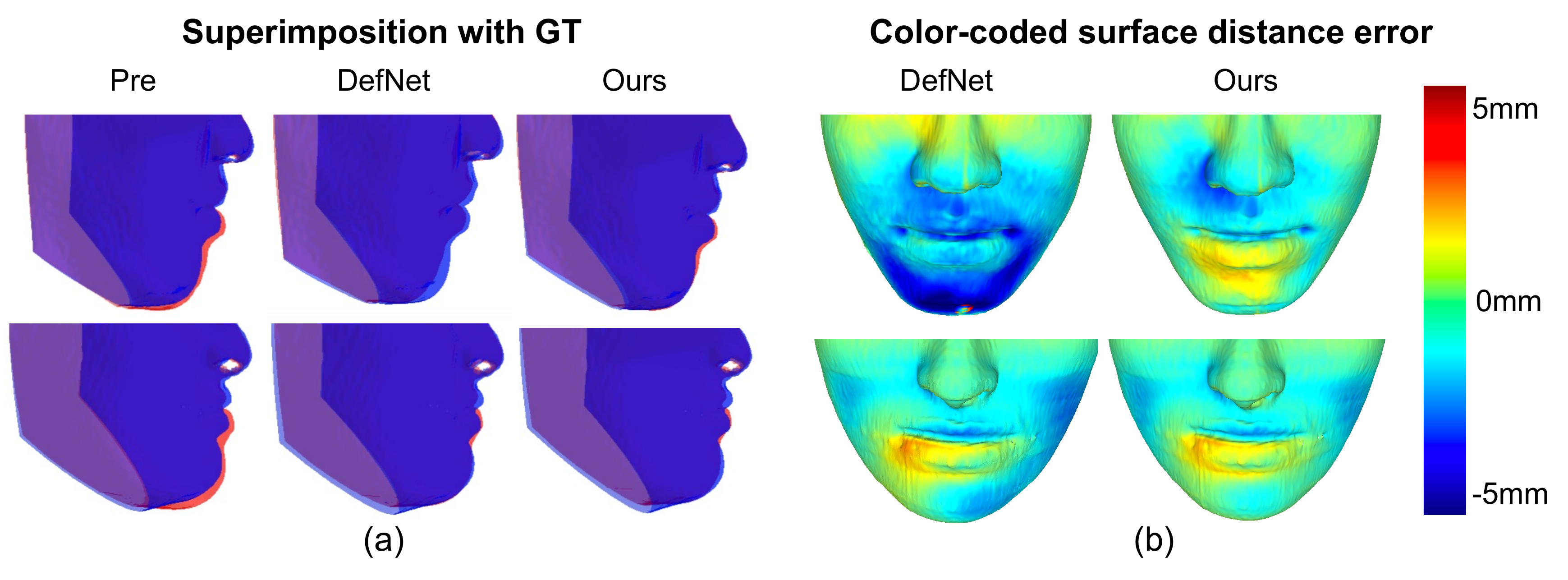}
	\caption{Examples of the results of simulated facial outcomes. (a) Comparison of the simulated facial outcome(red) with ground truth (blue). (b) Color-coded error map of the simulated facial outcome compared with ground truth.
	}
	\label{fig:facial}
\end{figure}

The evaluation of the bony surface showed that the proposed method successfully predicted bony surfaces that were similar to the real postoperative bones. 
To further assess the performance of the method, a facial simulator was used to qualitatively verify the simulated faces from the surgical plans. 
Fig~\ref{fig:facial} shows the comparison of the simulated facial outcomes of different methods using FS network and the desired facial appearance. 
The facial outcomes derived from bony plans of DefNet and our method are visualized, and the preoperative face is also superimposed with GT for reference.  
The results of the facial appearance simulation further validate the feasibility of the proposed learning-based framework for bony movement estimation and indicate that our method can achieve comparable simulation accuracy with the real bony plan.


\section{Discussions and Conclusions}
As a result of our approach that considers both the bony and soft-tissue components of the deformity, the accuracy of the estimated bony plan, especially on the DI segment, has significantly improved. 
Nonetheless, the non-linear relationship between the facial and bony surfaces cannot be adequately learned using only facial and bony surface data, and additional information such as tissue properties can also affect the facial outcome.
To account for uncertainty, we introduce random perturbations to generate different plans. 
In the future, we plan to incorporate additional uncertainty into the bony planner by using stronger perturbations or other strategies such as dropout\cite{gal2016dropout,Xu2022_TMI-SCOSSL} and adversarial attacks~\cite{zhang2022overlooked,Zhang2023aaai}, which could help create more diverse bony plans.
Also, relying solely on a deep learning-based facial simulation to evaluate our method might not fully validate its effectiveness.
We plan to utilize biomechanical models such as FEM to validate the efficacy of our approach in the future.
Moreover, for our ultimate goal of translating the approach to clinical settings, we will validate the proposed method using a larger patient dataset and compare the predicted bony plans with the actual surgical plans.

In conclusion, we have developed a soft-tissue driven framework to directly predict the bony plans that achieve a desired facial outcome.
Specifically, a bony planner and a facial simulator have been proposed for generating bony plans and verifying their effects on facial appearance. 
Evaluation results on a clinical dataset have shown that our method significantly outperforms the traditional bone-driven approach.  
By adopting this approach, we can create a virtual surgical plan that can be assessed and adjusted before the actual surgery, reducing the likelihood of complications and enhancing surgical outcomes.
The proposed soft-tissue driven framework can potentially improve the accuracy and efficiency of CMF surgery planning by automating the process and incorporating a facial simulator to account for the complex non-linear relationship between bony structure and facial soft-tissue.
\\
\\
\noindent \textbf{Acknowledgements.} 
This work was partially supported by NIH under awards R01 DE022676, R01 DE027251, and R01 DE021863.
%

\bibliographystyle{splncs04}
\bibliography{references}
\end{document}